\documentclass{article}

\usepackage[final]{neurips_2021}

\usepackage[utf8]{inputenc} 
\usepackage[T1]{fontenc}    
\usepackage{hyperref}       
\usepackage{url}            
\usepackage{booktabs}       
\usepackage{amsfonts}       
\usepackage{nicefrac}       
\usepackage{microtype}      
\usepackage{xcolor}         
\usepackage{comment}
\usepackage{enumitem}

\title{Whose Ground Truth? Accounting for Individual and Collective Identities Underlying Dataset Annotation
}

\author{%
  Emily Denton\thanks{Equal contribution; authors listed alphabetically.} \\
  Google Research \\
  \texttt{dentone@google.com} \\
   \And
   Mark Díaz\textsuperscript{$*$} \\
   Google Research\\
   \texttt{markdiaz@google.com} \\
  \And
   Ian Kivlichan\textsuperscript{$*$} \\
   Jigsaw \\
   \texttt{kivlichan@google.com} \\
    \And
   Vinodkumar Prabhakaran\textsuperscript{$*$} \\
   Google Research\\
   \texttt{vinodkpg@google.com} \\
   \And
   Rachel Rosen\textsuperscript{$*$} \\
   Jigsaw \\
   \texttt{rachelrosen@google.com} \\
}

\begin{document}

\maketitle

\begin{abstract}
Human annotations play a crucial role in machine learning (ML) research and development. However, the ethical considerations around the processes and decisions that go into building ML datasets has not received nearly enough attention. In this paper, we survey an array of literature that provides insights into ethical considerations around crowdsourced dataset annotation. We synthesize these insights, and lay out the challenges in this space along two layers: (1) who the annotator is, and how the annotators' lived experiences can impact their annotations, and (2) the relationship between the annotators and the crowdsourcing platforms and what that relationship affords them. Finally, we put forth a concrete set of recommendations and considerations for dataset developers at various stages of the ML data pipeline: task formulation, selection of annotators, platform and infrastructure choices, dataset analysis and evaluation, and dataset documentation and release.
\end{abstract}

\section{Introduction}
By enabling efficient and scalable distribution of data labelling microtasks, crowdsourcing platforms are a natural choice for dataset developers aiming to cheaply and efficiently generate dataset annotations.
In this short survey paper we explore the inherent  challenges and decision points that stem from crowdsourced dataset annotation. In particular, we ask: who is annotating the data, and why is that important? We consider how the ethical concerns of data annotation intersect with the identities of the annotators, the social structures surrounding their work, and how their individual perspectives may become encoded within the dataset labels. Data generated in crowdwork tasks is shaped by a range of social factors and the datasets that workers help to build continue to shape systems long after worker engagement ends. We argue that this impacts future models built from this data, and that understanding the perspectives captured within datasets is crucial to understanding resulting models and their potential social impact. 

Our work complements and extends prior scholarship examining ethical considerations relating to crowdsourcing \citep{vakharia2015beyond, schlagwein2019ethical, kocsis2016towards, shmueli2021beyond}. 
Our work is distinct from previous scholarship in that we focus our attention on unresolved ethical problems in crowdsourcing that relate specifically to individual worker subjectivity and individual worker experiences. 
We start by outlining a comprehensive set of concerns regarding how annotators' individual and collective social experiences, as well as their working conditions, may impact the nature of the data they provide for machine learning development, in particular the biases that may be captured in and propagated through those datasets. Based on this analysis, we offer a set of ethical considerations and recommendations for dataset developers that apply to different steps of a typical data annotation pipeline, from task formulation to dataset release.

\section{Who is annotating ML datasets and why does it matter?}
\label{sec_challenges}

Recent empirical work has revealed that relatively little attention is given or documented about annotator positionality---how annotator social identity shapes their understanding of the world. \citep{geiger2020, scheuerman2021}. Crowd workers are often selected by task requesters based on quality metrics, rather than on any socially defining features of their knowledge or experience. This is concerning, since crowd-sourced annotations are often used to build datasets capturing subjective phenomena such as sentiment and hate-speech, and hence crowd workers' values and subjective judgments shape the perspectives that machine learning models learn from in a manner that is wholly unaccounted for. Indeed, crowdsourcing platforms are often explicitly designed in a manner that positions crowdworkers as \textit{interchangeable} \citep{irani2013turkopticon}. 

\textbf{Accounting for the socio-cultural backgrounds} of dataset annotators is important for at least two reasons. First, subjective interpretations of a task can produce divergent annotations across different communities \citep{sen2015turkers}. As \citet{aroyo2015truth} argue, the notion of ``one truth'' in crowdsourcing responses is a myth; disagreement between annotators, which is often viewed as negative, can actually provide a valuable signal. Secondly, since many crowdsourced annotator pools are socio-demographically skewed, there are implications for which populations are represented in datasets as well as which populations face the challenges of crowdwork \citep{irani2013turkopticon, gray2019ghost}. Accounting for skews in annotator demographics is critical for contextualizing datasets and ensuring responsible downstream use. In short, there is value in acknowledging, and accounting for, worker's socio-cultural background---both from the perspective of data quality and societal impact.

\textbf{Accounting for lived experiences of annotators as expertise} may be of great utility in some cases. Just as substantive work experience lends valuable domain expertise for a given problem (e.g.\ annotation of medical imagery by a medical professional), lived experience with, and proximity to, a problem domain can provide a valuable source of expertise for dataset annotation. 
For example, women experience higher rates of sexual harassment online compared to men, and among those who have experienced online abuse, women are more likely to identify it as such \citep{vogels2021state}.
However, such lived experiences do not always fall along demographic lines. \cite{waseem2016racist} demonstrated that incorporating feminist and antiracist activists' perspectives into hate speech annotations yielded better aligned models. Similarly, \citet{patton2019annotating} demonstrated the importance of situated domain expertise --- including contextualized knowledge of local language, concepts, and gang activity --- when annotating Twitter images to detect pathways to violence among gang-involved youth in Chicago.

In summary, a core question to answer in data collection is how much annotator subjectivity matters for the task at hand, and how it impacts what the resulting dataset is meant to capture. While we used relatively subjective tasks as examples above, even seemingly objective tasks such as annotating medical texts vary surprisingly with annotator backgrounds and experience \citep{aroyo2015truth}.

\section{Worker Experiences of Dataset Annotation}

Another layer of considerations relate to annotators' experiences with annotation work itself and how it can impact how they do their work. These include issues related to worker compensation, imbalances in the relationship between worker and requester, and the structure of annotation work itself --- all of which can pose barriers to crowdworker well-being and their ability to produce quality work.

\textbf{Compensation policies} of the platforms should be a core aspect to consider when thinking about responsible data collection. For instance, in the U.S., there are currently no regulations around worker pay for crowdwork \citep{berg2015income}, and the Fair Labor Standards Act that established the minimum wage,\footnote{https://www.dol.gov/agencies/whd/flsa} is not applicable for crowdworkers as they are independent contractors \citep{semuels2018internet}. Moreover, for every hour of paid work, workers spend another 18 minutes on unpaid work, including searching for tasks \citep{berg2015income}. Time spent working is compounded by competition from other crowdworkers \citep{semuels2018internet}, which can pressure workers to be constantly available to look for work as well as work longer hours \citep{berg2015income}. In addition, a large majority of crowdworkers (94\% as per \citep{berg2015income}) have had work that was rejected or for which they were not paid. Yet, requesters retain full rights over the data they receive regardless of whether they accept or reject it; \citet{roberts2016digital} describes this system as one that ``enables wage theft''. Moreover, rejecting work and withholding pay is painful because rejections are often caused by unclear instructions and the lack of meaningful feedback channels; many crowdworkers report that poor communication negatively affects their work \citep{berg2015income}.

\textbf{Power dynamics} between the requesters and annotators is another major challenge. Top-down organizational structures often results in the workers viewing requesters as more informed as they are the ones who provided the data and the label schema \citep{miceli2020between}. Hence, instead of resolving ambiguities, workers are more likely to try to judge from the standpoint of the requester, often with limited exposure to the goals of the annotation. This contributes to the \textit{portability trap} \citep{selbst2018fairness}: a ``failure to understand how repurposing algorithmic solutions designed for one social context may be misleading, inaccurate, or otherwise do harm when applied to a different context.''
Power asymmetries also reflect global power dynamics. For instance, since technology development happens primarily in the West, human computation from the Global South is often relegated to the margins \citep{sambasivan2021re}.  

In summary, a core consideration for responsible data collection is whether there exist mechanisms for the workers to address these power asymmetries. The anonymous and geographically distributed nature of crowdsourced annotation work imposes significant barriers to collective action. In response, several community forums have been developed independently from crowdwork platforms to support crowd workers, e.g.\ TurkerNation, Turk Alert, MTurkGrind, and Reddit’s /r/HITsWorthTurkingFor. Turkopticon \citep{irani2013turkopticon, turkopticon} and Dynamo \citep{Salehi2015} have also emerged as activist tools that support and enable collective action for crowdworkers.

\section{Implications for Dataset Developers}

We now outline a comprehensive set of considerations for the collection, use, and dissemination of crowd-sourced ML datasets. We discuss them as they apply to different parts of a typical dataset construction pipeline, from the formulation of tasks to dissemination of datasets. 

\textbf{Task formulation:} A core objective of constructing ML datasets is to capture the aspects of human intelligence that are of importance to a given task. While some tasks tend to pose objective questions with a correct answer (\textit{is there a human face in an image?}), oftentimes datasets aim to capture judgement on relatively subjective tasks with no universally correct answer (\textit{is this piece of text offensive?}). It is important to be intentional about whether to lean on annotators' subjective judgements. This determination should be tied to the purpose of dataset creation and the downstream use cases it is meant to serve, rather than what is convenient, efficient, or scalable. Not accounting for task subjectivity may lead to inadvertent biases and misses critical insights about tasks that could benefit from the annotators' lived experiences. However, even for relatively subjective tasks such as labeling offensiveness, a dataset developer may want to restrict the annotators from relying on their lived experiences, say, if the dataset is meant to capture a set of policies defined by a platform. 
Clarifying such aspects of the tasks, has ramifications for how successfully the datasets capture the aspects of human intelligence they are meant to capture.

\textit{Recommendations and considerations}
\begin{itemize}[itemsep=1pt,topsep=0pt]
    \item Consider the subjective nature of your annotation task. Is it possible that individuals with different social and cultural backgrounds might differ in their judgements?
    \item Consider the forms of expertise that should be incorporated through data annotation, including both formal disciplinary training and lived experience with the problem domain. What are the risks of this expertise not being reflected in the annotator pool?
    \item 
    Make sure task instructions are clear and unambiguous in order to prevent annotators from wasting time on a task where their work will be rejected due to misunderstandings.
    \item Consider how the final dataset annotations will relate to individual annotator responses. For instance, will you release only the aggregated labels, e.g. through a majority vote?
    Consider what valuable information might be lost through such aggregation.
\end{itemize}

\textbf{Selecting annotators:}
As outlined in Section~\ref{sec_challenges}, the selection of an annotator pool is a highly consequential decision, especially given the subjective nature of many annotation tasks. 
It is important to choose annotation platforms that allow flexibility in designing custom annotator pools along various socio-demographic axes. These decisions should ideally be guided by considering which communities will be most impacted by models built from the data, and which communities could be harmed the most if they are not represented in the annotator pool.

\textit{Recommendations and considerations}
\begin{itemize}[itemsep=1pt,topsep=0pt]
    \item While there is no single ``correct'' way to assemble an annotator pool, 
    the decisions in this stage could impact the biases captured in the resulting dataset. For instance, annotator demographics may serve as a form of expertise that is important for the task (cf.\ Section~\ref{sec_challenges}).
    \item Consider the intended usage contexts of the dataset, and the marginalized communities therein, when choosing which annotators to be prioritized to be included. 
    \item Consider how labor practices intersect with the choice of who the annotators are. For example: if female annotators make up the majority as they do in the U.S.\ \citep{posch2018characterizing}, consider how fair payment, or the lack thereof, could impact this group.
\end{itemize}

\textbf{Platform and infrastructure choices:}
As described in Section~\ref{sec_challenges}, the platform policies around compensation and power asymmetries play a huge role in the quality of work the annotators produce. 
Some platforms offer platform-mediated channels of communication that allow task requesters to incorporate annotator feedback into the task framing or annotator guidelines. 
Different platforms also impose different minimum-pay constraints; requesters may want to support platforms that uphold fair pay standards. Separately from the platform, task creators should be aware of worker pay per hour, since this number is not often given explicitly. Some platforms may only offer requesters the option to select pay per item for an annotation task, and the defaults may be set low: task creators should take care when estimating work time per item to ensure they are paying workers fairly.

\textit{Recommendations and considerations}
\begin{itemize}[itemsep=1pt,topsep=0pt]
    \item Consider platform's underlying annotator pool and the options they provide to source specialized rater pools, and whether they enable you to curate an appropriate pool of annotators (e.g.\ considering sociodemographic factors or domain expertise). 
    \item Consider comparing and contrasting the minimum pay requirements established across different platforms. You may choose to support a platform that upholds fair pay standards.  
    \item Consider the extent to which you would like to establish a channel of communication and feedback between your team and the annotators. Platform mediated channels of communication can give annotators an opportunity to provide feedback on confusing instructions.
\end{itemize}

\textbf{Dataset analysis and evaluation:}
A common practice in building crowdsourced annotations is to obtain multiple annotator judgements that are then aggregated (e.g.,\ through majority voting) to obtain a single ``ground truth'' that is released in the dataset \citep{sabou2014corpus}. However, the disagreements between annotators may embed valuable nuances about the task \citep{alm2011subjective,aroyo2013crowd}. Aggregation, in such cases may obscure such nuances, and in that process potentially exclude perspectives from minority annotators \citep{prabhakaran2021releasing}.

\textit{Recommendations and considerations}
\begin{itemize}[itemsep=1pt,topsep=0pt]
    \item Consider including the uncertainty or disagreement between individual annotations on each instance as a signal in the dataset.
    \item Consider analyzing systematic disagreements between annotators of different socio-demographic groups in order to better understand how diverse perspectives are represented.
\end{itemize}

\textbf{Dataset documentation and release:}
Rigorous documentation of design decisions and outcomes relating to dataset annotations is an important aspect of responsible dataset development. Several dataset documentation frameworks have been proposed to contextualize a dataset and offer guidance regarding intended or unintended use, as well as facilitate accountability for development decisions.

\textit{Recommendations and considerations}
\begin{itemize}[itemsep=1pt,topsep=0pt]
    \item Consider adopting or adapting an existing dataset documentation framework (e.g.\ \cite{gebruDatasheetsDatasets2020, hollandDatasetNutritionLabel2018, benderDataStatementsNatural2018, kazimzade2020biased, Hutchinson2021}) to guide your dataset documentation. Consider publishing aggregate statistics on the sociodemographic make-up of your annotator pool.
    \item Consider including individual annotator responses for each data points in the dataset in addition to a final aggregated ground truth label, where applicable.
\end{itemize}

\bibliographystyle{ACM-Reference-Format}
\bibliography{main}

\end{document}